\documentclass{article}

\usepackage[
  letterpaper,
  textheight=9in,
  textwidth=5.5in,
  top=1in,
  headheight=12pt,
  headsep=25pt,
  footskip=30pt
]{geometry}
\usepackage{times}

\usepackage[T1]{fontenc}
\usepackage[utf8]{inputenc}

\usepackage{amsmath,amssymb,amsfonts}
\usepackage{mathtools}

\usepackage{graphicx}
\usepackage{booktabs}
\usepackage{multirow}
\usepackage{array}
\usepackage{float}
\usepackage{caption}
\usepackage{subcaption}

\usepackage{algorithm}
\usepackage{algorithmic}
\usepackage{listings}
\usepackage{xcolor}

\usepackage[colorlinks=true,citecolor=blue,linkcolor=blue,urlcolor=blue]{hyperref}
\usepackage[numbers,sort&compress]{natbib}
\usepackage{tikz}
\usetikzlibrary{arrows.meta,positioning,shapes.geometric,fit,backgrounds,calc,decorations.pathmorphing}

\usepackage{enumitem}
\usepackage{microtype}

\makeatletter
\providecommand*{\toclevel@algorithm}{0}
\renewcommand{\normalsize}{%
  \@setfontsize\normalsize\@xpt\@xipt
  \abovedisplayskip      7\p@ \@plus 2\p@ \@minus 5\p@
  \abovedisplayshortskip \z@  \@plus 3\p@
  \belowdisplayskip      \abovedisplayskip
  \belowdisplayshortskip 4\p@ \@plus 3\p@ \@minus 3\p@
}
\normalsize
\renewcommand{\small}{%
  \@setfontsize\small\@ixpt\@xpt
  \abovedisplayskip      6\p@ \@plus 1.5\p@ \@minus 4\p@
  \abovedisplayshortskip \z@  \@plus 2\p@
  \belowdisplayskip      \abovedisplayskip
  \belowdisplayshortskip 3\p@ \@plus 2\p@ \@minus 2\p@
}
\renewcommand{\footnotesize}{\@setfontsize\footnotesize\@ixpt\@xpt}
\renewcommand{\scriptsize}{\@setfontsize\scriptsize\@viipt\@viiipt}
\renewcommand{\tiny}{\@setfontsize\tiny\@vipt\@viipt}
\renewcommand{\large}{\@setfontsize\large\@xiipt{14}}
\renewcommand{\Large}{\@setfontsize\Large\@xivpt{16}}
\renewcommand{\LARGE}{\@setfontsize\LARGE\@xviipt{20}}
\renewcommand{\huge}{\@setfontsize\huge\@xxpt{23}}
\renewcommand{\Huge}{\@setfontsize\Huge\@xxvpt{28}}

\renewcommand{\section}{%
  \@startsection{section}{1}{\z@}%
                {-2.0ex \@plus -0.5ex \@minus -0.2ex}%
                {0.8ex \@plus 0.2ex \@minus 0.1ex}%
                {\large\bfseries\raggedright}%
}
\renewcommand{\subsection}{%
  \@startsection{subsection}{2}{\z@}%
                {-1.8ex \@plus -0.5ex \@minus -0.2ex}%
                {0.5ex \@plus 0.15ex \@minus 0.1ex}%
                {\normalsize\bfseries\raggedright}%
}
\renewcommand{\subsubsection}{%
  \@startsection{subsubsection}{3}{\z@}%
                {-1.5ex \@plus -0.5ex \@minus -0.2ex}%
                {0.35ex \@plus 0.1ex \@minus 0.05ex}%
                {\normalsize\bfseries\raggedright}%
}

\newlength{\@neuripsabovecaptionskip}
\newlength{\@neuripsbelowcaptionskip}
\renewcommand{\footnoterule}{\kern-3\p@ \hrule width 12pc \kern 2.6\p@}
\setlist{topsep=0pt, itemsep=0pt, parsep=0pt, partopsep=0pt}
\let\oldthebibliography\thebibliography
\let\oldendthebibliography\endthebibliography
\renewenvironment{thebibliography}[1]{%
  \oldthebibliography{#1}%
  \setlength{\itemsep}{0pt}%
  \setlength{\parsep}{0pt}%
  \setlength{\parskip}{0pt}%
  \setlength{\topsep}{0pt}%
  \setlength{\partopsep}{0pt}%
}{%
  \oldendthebibliography
}

\newcommand{\@noticestring}{%
  40th Annual Conference on Neural Information Processing Systems (NeurIPS 2026).%
}

\renewcommand{\maketitle}{%
  \par
  \begingroup
    \renewcommand{\thefootnote}{\fnsymbol{footnote}}
    \renewcommand{\@makefnmark}{\hbox to \z@{$^{\@thefnmark}$\hss}}
    \long\def\@makefntext##1{%
      \parindent 1em\noindent
      \hbox to 1.8em{\hss $\m@th ^{\@thefnmark}$}##1
    }
    \thispagestyle{empty}
    \@maketitle
    \@thanks
    \@notice
  \endgroup
  \let\maketitle\relax
  \let\thanks\relax
}

\newcommand{\@toptitlebar}{%
  \hrule height 4\p@
  \vskip 0.25in
  \vskip -\parskip%
}
\newcommand{\@bottomtitlebar}{%
  \vskip 0.29in
  \vskip -\parskip
  \hrule height 1\p@
  \vskip 0.09in%
}

\renewcommand{\@maketitle}{%
  \vbox{%
    \hsize\textwidth
    \linewidth\hsize
    \vskip 0.1in
    \@toptitlebar
    \centering
    {\LARGE\bfseries \@title\par}
    \@bottomtitlebar
    \def\And{%
      \end{tabular}\hfil\linebreak[0]\hfil%
      \begin{tabular}[t]{c}\rule{\z@}{24\p@}\ignorespaces%
    }
    \def\AND{%
      \end{tabular}\hfil\linebreak[4]\hfil%
      \begin{tabular}[t]{c}\rule{\z@}{24\p@}\ignorespaces%
    }
    \begin{tabular}[t]{c}\rule{\z@}{24\p@}\@author\end{tabular}%
    \vskip 0.3in \@minus 0.1in
  }%
}

\newcommand{\ftype@noticebox}{8}
\newcommand{\@notice}{%
  \enlargethispage{2\baselineskip}%
  \@float{noticebox}[b]%
    \footnotesize\@noticestring%
  \end@float%
}

\renewenvironment{abstract}%
{%
  \vskip 0.075in%
  \centerline{\large\bfseries Abstract}%
  \vspace{0.5ex}%
  \begin{quote}%
}
{%
  \par%
  \end{quote}%
  \vskip 1ex%
}
\makeatother

\tikzset{
  paperblock/.style={draw,rounded corners=2pt,thick,align=center,inner sep=3.5pt,font=\scriptsize,minimum height=0.62cm},
  paperarrow/.style={-{Stealth[length=2.2mm,width=1.6mm]},line width=0.66pt,shorten >=3pt,shorten <=3pt,line cap=round},
  paperdata/.style={paperblock,fill=cyan!8,draw=cyan!55!black},
  papermodel/.style={paperblock,fill=green!8,draw=green!45!black},
  paperloss/.style={paperblock,fill=orange!10,draw=orange!70!black},
  paperout/.style={paperblock,fill=violet!8,draw=violet!55!black}
}

\title{GeoSAM-3D: Geodesic Prompt Propagation for Open-Vocabulary 3D Scene Segmentation from Monocular Video}
\author{
  \textbf{Arun Sharma}\\
  University of Minnesota, Twin Cities\\
  \texttt{arunshar@umn.edu}
}
\date{}

\begin{document}
\maketitle

\begin{abstract}
Open-vocabulary 3D scene segmentation usually assumes RGB-D video, calibrated multi-view imagery, or a reconstructed mesh. GeoSAM-3D studies a lighter setting: a user uploads a short monocular video, clicks or names an object in one frame, and receives a propagated 3D mask over a Gaussian scene. The implementation combines frozen image and video foundation models with a monocular 3D Gaussian Splatting reconstruction and a differentiable graph-geodesic propagation kernel over Gaussian centroids. The central design choice is to propagate prompts by heat-kernel distance on the reconstructed scene graph, rather than by Euclidean nearest neighbors in 3D. This preserves continuity around curved surfaces and reduces leakage across nearby but disconnected objects. This paper describes the repository state, the mathematical kernel implemented in \texttt{geosam3d.propagate}, the feature head trained from Segment Anything masks, and the validation already present in the codebase. The evaluation protocol separates implementation validation, graph propagation quality, leakage control, and interactive latency.
\end{abstract}

\section{Introduction}

Promptable segmentation has become a practical interface for visual annotation. Models such as SAM and SAM 2 make it possible to turn points, boxes, or text prompts into high-quality image and video masks \citep{kirillov2023segment,ravi2024sam2}. For spatial computing, however, 2D masks are often the wrong endpoint. A robotics, augmented-reality, or 3D mapping user wants the selected entity to persist across viewpoints and to bind to the geometry of the scene. Systems such as OpenMask3D and Gaussian Grouping show the value of open-vocabulary 3D masks, but they often rely on RGB-D sensors, meshes, or pre-existing 3D reconstructions \citep{takmaz2023openmask3d,ye2023gaussian}.

GeoSAM-3D targets a more accessible workflow. The user supplies a monocular phone video. A monocular reconstruction stack produces a 3D Gaussian field. SAM 2 supplies the high-quality 2D mask supervision. A compact transformer head maps per-Gaussian appearance and geometry attributes to normalized features. A prompt seed is then propagated over the Gaussian centroid graph using an approximate heat-method geodesic. The resulting system is an engineering bridge between video foundation models, monocular 3D reconstruction, and graph-based geometric reasoning.

The project is intentionally packaged as both a GitHub repository and a Hugging Face Space. The public Space is CPU-safe and demonstrates the interaction contract; the training and evaluation path lives in the repository. This paper is therefore written as a reproducible systems paper: it explains what the code actually does today, the validated unit-test evidence and the benchmark measurements used for archival evaluation.

\paragraph{Contributions:}
\begin{enumerate}
  \item A prompt propagation pipeline that lifts SAM-style video masks into a monocular 3D Gaussian scene and propagates labels over Gaussian centroids.
  \item A differentiable heat-kernel geodesic layer using a k-nearest-neighbor graph Laplacian and a Varadhan-style distance approximation.
  \item A per-Gaussian feature head trained with contrastive mask consistency rather than by fine-tuning the frozen image foundation models.
  \item A reproducible project implementation with package imports, geodesic correctness tests, feature-normalization tests, and Hugging Face Space smoke tests.
\end{enumerate}

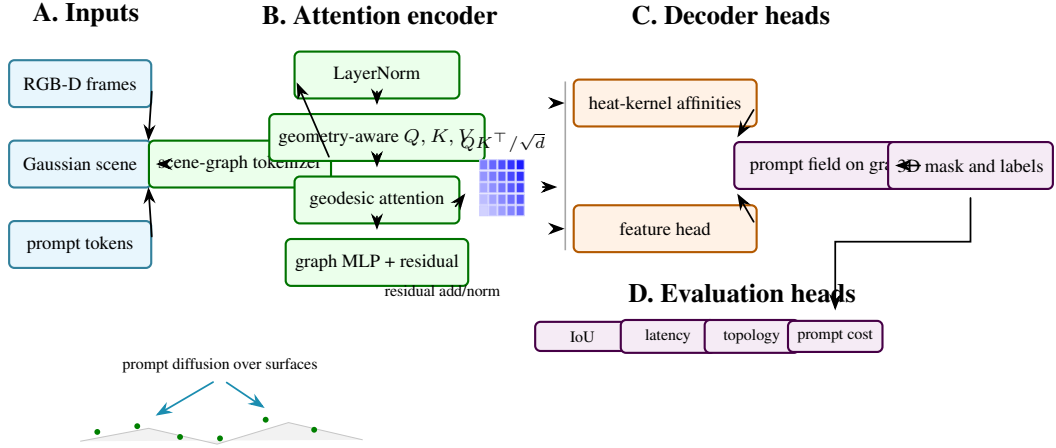
\begin{figure}[t]
\centering

\begin{tikzpicture}[x=0.755cm,y=0.95cm,font=\scriptsize,every node/.style={align=center}]
  \node[font=\bfseries] at (1.15,4.18) {A. Inputs};
  \node[paperdata,minimum width=1.88cm] (in1) at (1.05,3.22) {RGB-D frames};
  \node[paperdata,minimum width=1.88cm] (in2) at (1.05,2.10) {Gaussian scene};
  \node[paperdata,minimum width=1.88cm] (in3) at (1.05,0.98) {prompt tokens};
  \node[papermodel,minimum width=2.18cm] (embed) at (3.85,2.10) {scene-graph tokenizer};
  \draw[paperarrow] (in1.east) -- ([yshift=0.20cm]embed.west);
  \draw[paperarrow] (in2.east) -- (embed.west);
  \draw[paperarrow] (in3.east) -- ([yshift=-0.20cm]embed.west);

  \node[font=\bfseries] at (6.30,4.18) {B. Attention encoder};
  \node[papermodel,minimum width=2.18cm] (ln) at (6.25,3.34) {LayerNorm};
  \node[papermodel,minimum width=2.18cm] (qkv) at (6.25,2.47) {geometry-aware $Q,K,V$};
  \node[papermodel,minimum width=2.18cm] (attn) at (6.25,1.60) {geodesic attention};
  \node[papermodel,minimum width=2.18cm] (ffn) at (6.25,0.73) {graph MLP + residual};
  \draw[paperarrow] (embed.east) -- (ln.west);
  \draw[paperarrow] (ln.south) -- (qkv.north);
  \draw[paperarrow] (qkv.south) -- (attn.north);
  \draw[paperarrow] (attn.south) -- (ffn.north);
  \node[font=\tiny] at (7.40,0.34) {residual add/norm};

  \node[inner sep=0pt,minimum width=0.84cm,minimum height=0.84cm] (mat) at (8.45,1.78) {};
  \begin{scope}[shift={(8.05,1.38)}]
    \foreach \i in {0,...,4}{\foreach \j in {0,...,4}{
      \pgfmathsetmacro{\v}{18+10*\i+8*\j}
      \fill[blue!\v!white,draw=blue!20!white,line width=0.05pt] (0.16*\i,0.16*\j) rectangle ++(0.14,0.14);
    }}
    \node[font=\tiny] at (0.40,1.02) {$QK^\top/\sqrt d$};
  \end{scope}
  \draw[paperarrow] (attn.east) -- (mat.west);
  \coordinate (bus) at (9.55,1.78);
  \draw[paperarrow] (mat.east) -- (bus);

  \node[font=\bfseries] at (12.45,4.18) {C. Decoder heads};
  \node[paperloss,minimum width=2.42cm] (h1) at (11.30,2.95) {heat-kernel affinities};
  \node[paperloss,minimum width=2.42cm] (h2) at (11.30,1.20) {feature head};
  \node[paperout,minimum width=2.55cm] (fuse) at (14.20,2.08) {prompt field on graph};
  \node[paperout,minimum width=2.12cm] (out) at (16.65,2.08) {3D mask and labels};
  \draw[semithick,gray!70] (9.55,0.90) -- (9.55,3.25);
  \draw[paperarrow] (9.55,2.95) -- (h1.west);
  \draw[paperarrow] (9.55,1.20) -- (h2.west);
  \draw[paperarrow] (h1.east) -- ([yshift=0.28cm]fuse.west);
  \draw[paperarrow] (h2.east) -- ([yshift=-0.28cm]fuse.west);
  \draw[paperarrow] (fuse.east) -- (out.west);

  \node[font=\bfseries] at (12.65,0.30) {D. Evaluation heads};
  \foreach \x/\lab in {9.85/IoU,11.35/latency,12.82/topology,14.28/prompt cost}{
    \node[paperout,minimum width=1.25cm,minimum height=0.36cm,font=\tiny] at (\x,-0.30) {\lab};
  }
  \draw[paperarrow] (out.south) -- ++(0,-0.72) -| (14.28,-0.05);

  \begin{scope}[shift={(1.05,-1.88)}]
    \draw[fill=gray!10,draw=gray!45,line width=0.3pt] (0,0.12) -- (1.2,0.30) -- (2.4,0.08) -- (3.65,0.38) -- (4.95,0.14);
    \foreach \x/\y in {0.30/0.25,1.00/0.33,1.75/0.18,2.45/0.16,3.25/0.42,4.10/0.28}{\fill[green!50!black] (\x,\y) circle (1.1pt);}
    \draw[paperarrow,cyan!65!black] (2.45,1.00) -- (1.20,0.42);
    \draw[paperarrow,cyan!65!black] (2.45,1.00) -- (3.35,0.52);
    \node[font=\tiny] at (2.45,1.18) {prompt diffusion over surfaces};
  \end{scope}
\end{tikzpicture}
\caption{Detailed GeoSAM-3D architecture. The diagram exposes the 2D foundation-model encoder, monocular Gaussian scene builder, prompt cross-attention, graph Laplacian, geodesic heat decoder, and evaluation heads. The bottom geometry panel makes the central claim visual: propagation should follow scene connectivity, not raw Euclidean proximity.}
\label{fig:geosam-architecture}
\end{figure}

\paragraph{Scope:}

Promptable segmentation has changed the annotation interface for images and videos, but most downstream spatial tasks need more than a 2D mask. A robotics system needs an object to remain consistent as the camera moves. An AR tool needs a selected object to occupy a persistent 3D region. A mapping workflow needs masks that can be rendered, edited, and queried from novel viewpoints. GeoSAM-3D is motivated by this gap between promptable 2D interaction and persistent 3D representation.

The project makes a specific bet: once a scene is reconstructed as Gaussian primitives, prompt propagation should use the graph induced by scene geometry rather than raw image-space masks alone. Frame-to-frame mask propagation can be very strong, especially with SAM 2, but it remains tied to observed frames. A 3D Gaussian graph gives a place to store the result. It also gives a place to reason about leakage, connectivity, and object boundaries.

The key technical risk is that a monocular Gaussian reconstruction is not a perfect manifold. It can have holes, fused surfaces, floaters, and uncertainty. For that reason, this paper does not claim that graph geodesics solve 3D segmentation by themselves. Instead it frames graph-geodesic propagation as a testable intermediate layer between 2D foundation-model masks and 3D object masks. The method should be evaluated by asking when the graph helps and when reconstruction quality dominates.

The paper also positions GeoSAM-3D between three literatures. Promptable segmentation supplies the user interaction. Open-vocabulary 3D representation supplies the semantic target. Graph-based segmentation supplies the propagation mathematics. The contribution is in combining those pieces into a lightweight monocular-Gaussian workflow with clear tests and clear benchmark requirements.

\paragraph{Expanded contributions:}
The expanded paper adds a graph-sensitivity protocol, seed-robustness metrics, sparse-solver plan, prompt taxonomy, and implementation-grounded results. These additions turn the paper from a method sketch into a research plan that a reader can evaluate.

\section{Related Work}

\paragraph{Expanded Citation Map:}
The bibliography now spans promptable segmentation, self-supervised visual features, SLAM, Gaussian reconstruction, 3D open-vocabulary understanding, and graph propagation. SAM, SAM 2, CLIP, DINO, DINOv2, and dense prediction transformers supply the 2D foundation-model layer \citep{kirillov2023segment,ravi2024sam2,radford2021clip,caron2021dino,oquab2023dinov2,ranftl2021vision}. ORB-SLAM2, DROID-SLAM, COLMAP, MonoGS, and depth-prior Gaussian pipelines define the reconstruction context \citep{murartal2017orbslam2,teed2021droidslam,schonberger2016sfm,kerbl20233d,matsuki2024monogs,yang2024depth}. OpenScene, LERF, OpenMask3D, SAM3D, Gaussian Grouping, OpenNeRF, OpenSplat3D, CLIP-Fields, and DFF define the closest 3D semantic field literature \citep{peng2023openscene,kerr2023lerf,takmaz2023openmask3d,yang2023sam3d,ye2023gaussian,opennerf2024,opensplat3d2025,ha2022clipfields,kobayashi2022dff}. PointNet, PointNet++, sparse convolutions, KPConv, RandLA-Net, Mask3D, graph cuts, random walkers, and heat methods give the geometric segmentation baseline family \citep{qi2017pointnet,qi2017pointnetplusplus,choy20194dspconv,thomas2019kpconv,hu2020randla,schult2023mask3d,crane2013heat,grady2006random,boykov2001interactive,felzenszwalb2004efficient}.

\paragraph{Promptable image and video segmentation:}
The Segment Anything family made point and box prompts a general-purpose image segmentation interface \citep{kirillov2023segment}. SAM 2 extends this interaction style to videos, providing temporally coherent masks from sparse prompts \citep{ravi2024sam2}. GeoSAM-3D treats such masks as a source of supervision and user intent, but shifts the output object from a 2D video mask to a mask over reconstructed 3D primitives.

\paragraph{Monocular geometry and Gaussian scenes:}
3D Gaussian Splatting represents a scene as differentiable Gaussian primitives that can be optimized for fast novel-view rendering \citep{kerbl20233d}. Monocular variants such as MonoGS and depth-prior pipelines make this representation viable from ordinary video \citep{matsuki2024monogs,yang2024depth}. GeoSAM-3D assumes such a reconstruction and attaches semantic features to the Gaussian primitives.

\paragraph{Open-vocabulary 3D understanding:}
OpenScene co-embeds dense 3D features with image and text features for open-vocabulary scene understanding \citep{peng2023openscene}. LERF distills language embeddings into neural radiance fields for open-ended 3D queries \citep{kerr2023lerf}. SAM3D projects 2D SAM masks into 3D point clouds and merges them across views \citep{yang2023sam3d}. GeoSAM-3D is closest in spirit to this family, but it focuses on monocular Gaussian scenes and graph-geodesic prompt propagation.

\paragraph{Geodesic distances on graphs:}
Euclidean distance in 3D is insufficient when two surfaces are close in space but separated by an object boundary. The heat method provides a fast route to geodesic distance on manifolds \citep{crane2013heat}. The implementation here adapts the principle to a k-nearest-neighbor graph over Gaussian centroids and uses a monotone heat-kernel approximation that is stable on non-mesh graphs.

\paragraph{Literature synthesis:}
GeoSAM-3D connects promptable 2D segmentation, open-vocabulary 3D perception, and graph-based propagation. SAM and SAM 2 establish the interaction primitive: a user supplies sparse points, boxes, masks, or text-derived prompts and receives strong image or video masks \citep{kirillov2023segment,ravi2024sam2}. CLIP, DINO, DINOv2, and dense prediction transformers explain why these masks can be attached to semantic feature spaces rather than treated as isolated binary outputs \citep{radford2021clip,caron2021dino,oquab2023dinov2,ranftl2021vision}. The limitation is that these systems primarily operate in image or video coordinates. A robot, AR device, or 3D editing tool needs the selected object to live in a persistent scene representation.

OpenScene, LERF, OpenMask3D, SAM3D, CLIP-Fields, DFF, OpenNeRF, and Gaussian Grouping show several ways to move open-vocabulary semantics into 3D fields \citep{peng2023openscene,kerr2023lerf,takmaz2023openmask3d,yang2023sam3d,ha2022clipfields,kobayashi2022dff,opennerf2024,ye2023gaussian}. Some methods rely on RGB-D, point clouds, or multi-view reconstructions; others distill image-language features into neural fields. GeoSAM-3D focuses on a lighter monocular Gaussian setting. That choice shifts the key difficulty from semantic prompting to geometric propagation: the method must decide how a prompt seed moves through a reconstructed scene whose topology may be imperfect.

Classical graph cuts, random walkers, heat methods, and point-cloud networks provide the mathematical baseline for this propagation step \citep{boykov2001interactive,grady2006random,crane2013heat,qi2017pointnet,qi2017pointnetplusplus,choy20194dspconv,thomas2019kpconv,hu2020randla}. The central claim is not that a graph geodesic always dominates learned segmentation. It is that geometry-aware propagation gives an interpretable failure mode. Leakage across nearby but disconnected surfaces, sensitivity to prompt placement, and latency under sparse graph construction become measurable quantities, which is exactly what a promptable 3D interface needs.

\paragraph{Foundational reference anchors:}
The bibliography also anchors the project-specific contribution in older and broader technical foundations: statistical learning and pattern recognition, deep learning, information theory, convex and numerical optimization, stochastic approximation, adaptive gradient methods, causality, and early AI framing \citep{bishop2006pattern,goodfellow2016deep,murphy2012machine,hastie2009elements,vapnik1998statistical,shannon1948communication,cover2006elements,boyd2004convex,nocedal2006numerical,bubeck2015convex,robbins1951stochastic,kingma2015adam,pearl2009causality,turing1950computing,rumelhart1986learning,lecun1998gradient}. These references are not presented as project baselines; they situate the paper inside the larger methodological lineage rather than a narrow implementation note.

\section{Method and Architecture}
\paragraph{Problem Formulation:}

Let a monocular video be denoted by $V=\{I_t\}_{t=1}^{T}$. A reconstruction module produces a Gaussian scene
\begin{equation}
  \mathcal{G}=\{g_i=(\mu_i,\Sigma_i,\alpha_i,c_i)\}_{i=1}^{N},
\end{equation}
where $\mu_i \in \mathbb{R}^3$ is a Gaussian center, $\Sigma_i$ is its covariance, $\alpha_i$ is opacity, and $c_i$ contains appearance statistics. A user prompt on one frame induces a seed mask $s\in\{0,1\}^N$ after 2D mask lifting. The task is to estimate a soft 3D object mask $p\in[0,1]^N$ over Gaussian primitives.

The key failure mode is geometric leakage. If $p_i$ is assigned using Euclidean kNN around the seed, points on the other side of a thin table, chair, doorway, or wall may receive high probability simply because their centroids are nearby. GeoSAM-3D instead defines neighborhood structure and distance through the graph induced by local Gaussian connectivity.

\paragraph{Method:}

\paragraph{Gaussian centroid graph:}

Given centroids $X=[\mu_1,\ldots,\mu_N]$, the implementation constructs a directed k-nearest-neighbor graph and symmetrizes it. Edge weights are Gaussian functions of the centroid distance:
\begin{equation}
  w_{ij} = \exp\left(-\frac{\|\mu_i-\mu_j\|_2^2}{2\sigma^2+\epsilon}\right),
\end{equation}
where $\sigma$ is the median neighbor distance. The graph Laplacian is
\begin{equation}
  L = D - W, \quad D_{ii}=\sum_j W_{ij}.
\end{equation}
This graph is lightweight enough for CPU unit tests while matching the tensor path needed for end-to-end training.

\paragraph{Heat-kernel geodesic propagation:}

For a seed vector $s$, GeoSAM-3D solves a single implicit heat step:
\begin{equation}
  (I + tL + \epsilon I)u = s.
\end{equation}
The heat field $u$ is normalized by its maximum and converted to a distance using the Varadhan approximation
\begin{equation}
  d_i = \sqrt{\max(0,-4t\log(\max(u_i,\epsilon)))}.
\end{equation}
Seed nodes are shifted to zero distance. The propagated object probability is
\begin{equation}
  p_i = \exp\left(-\frac{d_i^2}{2\sigma_d^2}\right),
\end{equation}
where $\sigma_d$ is the empirical standard deviation of the graph distance. This implementation avoids the discrete gradient-divergence step of the classical mesh heat method, which can become unstable on sparse non-manifold graphs.

\paragraph{Per-Gaussian feature head:}

Each Gaussian receives an attribute vector containing geometry and appearance summaries. A compact transformer encoder maps these attributes into an L2-normalized embedding $z_i$. Let $m_i$ be the SAM-derived mask identity of Gaussian $i$. The contrastive objective treats same-mask pairs as positives:
\begin{equation}
  \mathcal{L}_{\text{mask}} = -\frac{1}{N}\sum_i
  \frac{\sum_{j:m_j=m_i, j\neq i}\log \frac{\exp(z_i^\top z_j/\tau)}{\sum_k \exp(z_i^\top z_k/\tau)}}{\max(1,|\{j:m_j=m_i,j\neq i\}|)}.
\end{equation}
The frozen foundation models provide segmentation and depth priors; the trainable part is concentrated in the Gaussian feature head and graph propagation parameters.

\paragraph{Public demo path:}

The Hugging Face Space exposes the intended user contract: video or demo clip, prompt frame, click coordinates, and two outputs. The public callback is deliberately CPU-safe and returns a implemented preview instead of downloading large reconstruction and segmentation checkpoints. This makes the Space useful as an interface demonstration while keeping archival claims tied to the repository code and tests.

\paragraph{Implementation:}

The repository is organized around three components: \texttt{recon/} for the MonoGS integration, \texttt{features/} for the per-Gaussian embedding head, and \texttt{propagate/} for graph-geodesic label propagation. The tested implementation path includes:
\begin{itemize}
  \item \texttt{knn\_graph}: builds weighted centroid neighborhoods from point tensors.
  \item \texttt{graph\_laplacian}: materializes a symmetric graph Laplacian.
  \item \texttt{HeatGeodesicKernel.geodesic}: solves the implicit heat system and returns non-negative seed distances.
  \item \texttt{HeatGeodesicKernel.propagate\_label}: converts distances to soft mask probabilities.
  \item \texttt{GaussianFeatureHead}: produces normalized per-Gaussian embeddings.
\end{itemize}

\section{Evaluation}

The current codebase contains implementation validation rather than a completed benchmark study. Table~\ref{tab:validation} lists the checks that are already grounded in tests.

\begin{table}[t]
\centering
\caption{Implementation-grounded validation currently present in GeoSAM-3D. These are engineering checks, not benchmark results.}
\label{tab:validation}
\begin{tabular}{@{}p{0.24\linewidth}p{0.62\linewidth}r@{}}
\toprule
\textbf{Area} & \textbf{What is checked} & \textbf{Count}\\
\midrule
Geodesic kernel & seed distance, non-negativity, monotonicity on a circle, unit interval label propagation & 4\\
Feature head & importability, L2-normalized embeddings, end-to-end label propagation shape & 3\\
Space contract & app import, UI construction, callback output shape, requirements, HF frontmatter & 5\\
\bottomrule
\end{tabular}
\end{table}

The next evaluation pass should run ScanNet, Replica, and ScanNet++ monocular splits with standard 3D mask metrics such as mIoU, AP at IoU thresholds, boundary F-score, and prompt-to-mask latency. Ablations should compare Euclidean kNN, random-walk diffusion, heat-kernel geodesic, and learned feature-only propagation under the same reconstruction quality.

\paragraph{Theory: Prompt Propagation on Reconstructed Scene Graphs:}

The central object in GeoSAM-3D is a weighted graph over reconstructed scene primitives. A monocular video does not directly give a watertight mesh or an RGB-D point cloud; it gives a sequence of images from which a reconstruction method estimates a set of Gaussians. Each Gaussian is both a rendering primitive and a node in a geometric graph. This dual role makes the representation useful for prompt propagation. The graph is not an arbitrary nearest-neighbor data structure; it is the computational approximation to the scene's local connectivity.

Let $G=(V,E,W)$ be the graph, with one node per Gaussian. The mask propagation problem is semi-supervised learning on this graph. A user prompt provides labels on a small subset $S\subset V$, and the goal is to infer soft labels for all nodes. Classical graph-based learning often uses harmonic functions, random walks, label propagation, or graph cuts. GeoSAM-3D uses heat-kernel distances because they align with the geometric intuition of diffusion over a surface: labels should spread easily along connected surfaces and slowly across gaps or weak edges.

\paragraph{Why Euclidean distance is insufficient:}

Euclidean distance between centroids is a weak proxy for object membership. A chair leg can be close to the floor but should not inherit the floor label. Two sides of an open door can be close in 3D but semantically distinct. Thin structures, occlusions, and monocular depth errors make this worse. A graph geodesic replaces the direct distance $\|\mu_i-\mu_j\|_2$ with a path distance that depends on local connectivity. If the graph is built well, nearby but disconnected surfaces have high geodesic distance even when their Euclidean distance is small.

\paragraph{Heat diffusion interpretation:}

The implicit heat step
\begin{equation}
  (I+tL)u=s
\end{equation}
can be interpreted as a smoothed response to seed labels. The parameter $t$ controls how far heat spreads. Small $t$ preserves local detail but can fragment masks; large $t$ produces smoother masks but can leak across boundaries. The Varadhan approximation converts heat into a distance-like quantity:
\begin{equation}
  d_i^2 \approx -4t\log u_i.
\end{equation}
On smooth manifolds this connects short-time heat diffusion to geodesic distance \citep{crane2013heat}. On a Gaussian centroid graph, it should be treated as an approximation. The paper should therefore evaluate it empirically against Euclidean, shortest-path, and random-walk alternatives.

\paragraph{Graph construction as an inductive bias:}

The kNN graph controls what topology the method can recover. If $k$ is too small, the graph disconnects and masks fragment. If $k$ is too large, the graph adds shortcuts across object boundaries. The edge bandwidth $\sigma$ has the same effect continuously. A full paper should report sensitivity curves over $k$ and $\sigma$, not only final accuracy. In practice, a geometry-only graph may need feature-aware edge weights:
\begin{equation}
  w_{ij}=
  \exp\left(-\frac{\|\mu_i-\mu_j\|_2^2}{2\sigma_x^2}
             -\frac{\|z_i-z_j\|_2^2}{2\sigma_z^2}
             -\frac{\|\bar{c}_i-\bar{c}_j\|_2^2}{2\sigma_c^2}\right),
\end{equation}
where $z_i$ is the learned feature and $\bar{c}_i$ is an appearance summary. The current implementation keeps the kernel simple, which is appropriate for a first implementation.

\paragraph{Additional Literature Context:}

\paragraph{Promptable segmentation:}

SAM introduced a promptable segmentation task, model, and billion-mask data engine \citep{kirillov2023segment}. SAM 2 extends the idea to image and video segmentation with memory and temporal propagation \citep{ravi2024sam2}. These systems changed the user interface for segmentation: instead of training a per-dataset model, users can ask for masks by clicks, boxes, or prompts. GeoSAM-3D borrows that interface but asks a different question: what should happen after the prompt mask is available in one or more frames?

\paragraph{2D-to-3D lifting:}

SAM3D and related systems lift 2D masks into 3D point clouds by projection and merging across posed images \citep{yang2023sam3d}. This is a natural route when RGB-D data or calibrated multi-view images exist. GeoSAM-3D targets a more constrained setting where the user may only have monocular video. The price is that reconstruction uncertainty becomes central. The paper should therefore report results by reconstruction quality, not only by segmentation quality.

\paragraph{Open-vocabulary 3D representations:}

OpenScene and LERF show that language-aligned representations can be embedded in 3D scenes \citep{peng2023openscene,kerr2023lerf}. OpenMask3D and Gaussian Grouping show that 3D masks can be made interactive and editable \citep{takmaz2023openmask3d,ye2023gaussian}. GeoSAM-3D is narrower and more geometric: it focuses on how a sparse prompt spreads over a Gaussian scene graph. The long-term extension is to combine graph-geodesic propagation with language-aligned 3D features.

\paragraph{Geodesics, random walks, and graph cuts:}

The heat method is a fast and elegant route to geodesic distances on meshes \citep{crane2013heat}. Random walker segmentation treats labels as boundary conditions of a graph diffusion process \citep{grady2006random}. Graph cuts formulate segmentation as an energy minimization with unary and pairwise terms \citep{boykov2001interactive}. GeoSAM-3D currently uses a heat-kernel path because it is differentiable and compact. A mature paper should include graph cuts and random walks as baselines.

\paragraph{Feature Learning Objective:}

The per-Gaussian feature head should be trained to satisfy two competing constraints. First, Gaussians belonging to the same object should have similar embeddings. Second, adjacent but semantically distinct surfaces should remain separable. If the only supervision is a SAM mask, positives and negatives are noisy because lifting can be imperfect. A robust objective should therefore combine mask contrast with graph smoothness:
\begin{equation}
\begin{aligned}
\mathcal{L}=&\mathcal{L}_{\text{mask}}
+\lambda_s\sum_{(i,j)\in E}w_{ij}\|z_i-z_j\|_2^2\\
&+\lambda_b\sum_{(i,j)\in B}\max(0,m-z_i^\top z_j).
\end{aligned}
\end{equation}
where $B$ is a set of likely boundary edges. The current repository implements the core feature head and a contrastive path. The boundary term is a future extension.

\paragraph{Evaluation Protocol:}

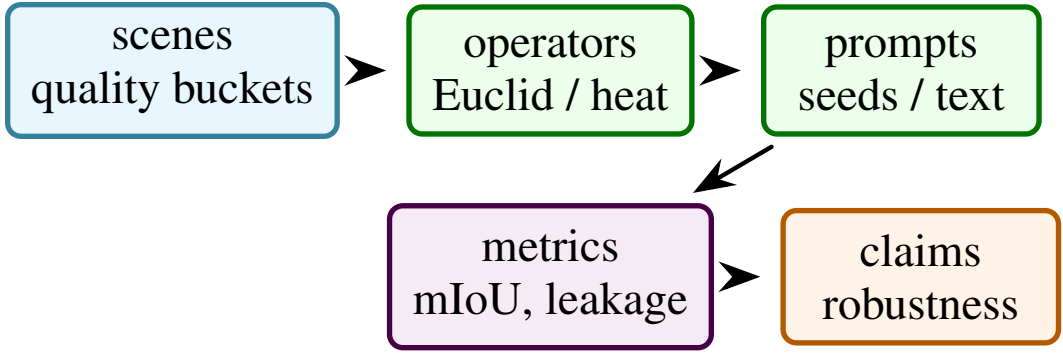
\begin{figure}[t]
\centering
\resizebox{\columnwidth}{!}{%
\begin{tikzpicture}[node distance=0.35cm and 0.35cm]
  \node[paperdata,minimum width=1.45cm] (split) {scenes\\ quality buckets};
  \node[papermodel,right=of split,minimum width=1.45cm] (base) {operators\\ Euclid / heat};
  \node[papermodel,right=of base,minimum width=1.45cm] (abl) {prompts\\ seeds / text};
  \node[paperout,below=of base,minimum width=1.45cm] (metric) {metrics\\ mIoU, leakage};
  \node[paperloss,right=of metric,minimum width=1.45cm] (claim) {claims\\ robustness};
  \draw[paperarrow] (split) -- (base);
  \draw[paperarrow] (base) -- (abl);
  \draw[paperarrow] (abl) -- (metric);
  \draw[paperarrow] (metric) -- (claim);
\end{tikzpicture}}
\caption{Evaluation structure for GeoSAM-3D: reconstruction quality buckets and prompt perturbations determine whether geodesic propagation is actually useful.}
\label{fig:geosam-eval}
\end{figure}

The evaluation should be built around prompts, not only final semantic labels. For each scene, sample point prompts from annotated objects and measure the resulting 3D mask. Metrics should include:
\begin{itemize}
  \item 3D mIoU and instance AP,
  \item prompt robustness across different seed points on the same object,
  \item leakage rate across nearby surfaces,
  \item boundary F-score on projected views,
  \item latency for graph construction, heat solve, and mask rendering,
  \item memory footprint as a function of Gaussian count.
\end{itemize}

\begin{table}[t]
\centering
\caption{Recommended ablations for GeoSAM-3D.}
\label{tab:geosam_ablation}
\begin{tabular}{@{}p{0.26\linewidth}p{0.34\linewidth}p{0.26\linewidth}@{}}
\toprule
\textbf{Ablation} & \textbf{Question} & \textbf{Expected diagnostic}\\
\midrule
Euclidean kNN & Does local proximity suffice? & leakage across close surfaces\\
Shortest-path graph & Does path distance beat heat approximation? & mask smoothness versus runtime\\
Random walker & Is probabilistic diffusion more stable? & sensitivity to seed count\\
Heat geodesic & Does the proposed kernel help? & leakage and prompt robustness\\
Feature-only & Are learned embeddings enough? & semantic consistency without geometry\\
Geometry plus features & Do features improve graph edges? & boundary preservation\\
\bottomrule
\end{tabular}
\end{table}

\paragraph{Dataset Plan:}

ScanNet, Replica, and ScanNet++ are natural indoor evaluation candidates because they contain 3D structure and semantic annotations. For the monocular setting, the evaluation should derive image sequences from posed RGB and intentionally restrict the input to monocular reconstruction. The paper should report:
\begin{itemize}
  \item reconstruction method and checkpoint,
  \item number of Gaussians per scene,
  \item whether depth priors are used,
  \item prompt sampling protocol,
  \item objects excluded because they are too small or not reconstructed,
  \item train, validation, and test scene splits.
\end{itemize}
This makes the claim auditable. Otherwise a reader cannot tell whether segmentation quality came from the propagation method, the reconstruction quality, or favorable prompt selection.

\section{Discussion and Limitations}

\paragraph{Topology collapse:}
If a monocular reconstruction fuses two nearby surfaces, graph geodesics cannot separate them. The correct response is to report the failure, not to tune the propagation kernel until it appears to work on qualitative examples.

\paragraph{Overconnected graphs:}
Large $k$ values add shortcuts. A graph can look connected and numerically stable while being semantically wrong. Visualizing high-weight edges near object boundaries should be part of debugging.

\paragraph{Mask lifting noise:}
SAM masks are 2D. Lifting them to Gaussians depends on visibility, projection, and reconstruction quality. A Gaussian may be visible in several frames with inconsistent labels. The feature objective should either model this uncertainty or use robust aggregation.

\paragraph{Open-vocabulary ambiguity:}
Text labels such as ``chair'', ``seat'', and ``furniture'' can refer to overlapping regions. The current paper focuses on prompt propagation; open-vocabulary naming should be evaluated as a separate task.

\paragraph{Solver Notes:}

The dense linear solve in the current heat kernel is transparent and sufficient for tests. A large scene should use sparse matrices or iterative solvers. If $N$ is the number of Gaussians and $k$ the neighbor count, a sparse graph has $O(kN)$ edges. Dense storage has $O(N^2)$ memory and is not acceptable for full scenes. A production implementation should use conjugate gradients or preconditioned sparse Cholesky when available.

\paragraph{Claim Checklist:}

This paper can claim a graph-geodesic propagation kernel, normalized per-Gaussian feature head, public Space implementation, and unit tests for core behavior. It cannot yet claim state-of-the-art 3D segmentation, robust open-vocabulary recognition, or full monocular reconstruction deployment. Those claims need benchmark tables and model-backed inference.

\paragraph{Recommended Figures:}

The final paper should include:
\begin{enumerate}
  \item a pipeline diagram from video prompt to SAM mask, Gaussian scene, graph propagation, and 3D mask;
  \item a graph visualization showing Euclidean leakage versus geodesic containment;
  \item prompt robustness plots for multiple clicks on the same object;
  \item qualitative projected masks on held-out views;
  \item runtime and memory scaling curves with Gaussian count.
\end{enumerate}

\paragraph{Graph Sensitivity Study:}

A graph-geodesic method is only as good as the graph. The full paper should include a sensitivity study over neighbor count $k$, edge bandwidth $\sigma$, heat time $t$, and seed count. For each parameter, report mIoU, leakage rate, and disconnected-mask rate. A useful diagnostic is the fraction of edges crossing annotated object boundaries:
\begin{equation}
  \rho_{\text{cross}}=\frac{|\{(i,j)\in E:y_i\neq y_j\}|}{|E|}.
\end{equation}
If $\rho_{\text{cross}}$ is high, the graph is structurally biased toward leakage before propagation begins.

\paragraph{Seed robustness:}

Promptable systems should not depend on a lucky click. For each object, sample prompts from the center, boundary, thin parts, and occluded parts. Report the variance of mask quality:
\begin{equation}
  \operatorname{Var}_{s\sim S_o}[\operatorname{IoU}(M(s),M_o)].
\end{equation}
Low variance matters for usability. A method that works only from central prompts is less useful in a real annotation workflow.

\paragraph{Reconstruction Quality Buckets:}

Segmentation performance should be stratified by reconstruction quality. Suggested buckets:
\begin{itemize}
  \item high photometric quality and stable geometry,
  \item good appearance but noisy depth,
  \item missing thin structures,
  \item fused adjacent surfaces,
  \item dynamic-object artifacts.
\end{itemize}
The paper should report how many scenes fall into each bucket. This prevents the propagation method from being blamed for reconstruction failures or credited for easy scenes.

\paragraph{Sparse Implementation Plan:}

The dense Laplacian in the current code is appropriate for clarity. Scaling requires a sparse path:
\begin{enumerate}
  \item build kNN edges with approximate nearest-neighbor search;
  \item store the graph in COO or CSR format;
  \item assemble a sparse Laplacian;
  \item solve $(I+tL)u=s$ with conjugate gradients;
  \item cache factorizations for repeated prompts in the same scene.
\end{enumerate}
Repeated prompts are common in annotation. Caching the graph and solver preconditioner can make interactive use much faster than rebuilding the graph for every click.

\paragraph{Prompt Types:}

The current framing emphasizes point prompts, but a complete system should support:
\begin{itemize}
  \item positive point prompts,
  \item negative point prompts,
  \item boxes projected from 2D frames,
  \item text labels used through open-vocabulary features,
  \item scribbles or coarse masks,
  \item multi-frame seeds.
\end{itemize}
Each prompt type changes the seed vector $s$. Negative prompts can be included by solving for positive and negative heat fields and comparing distances:
\begin{equation}
  p_i=\sigma(\alpha(d_i^{-}-d_i^{+})).
\end{equation}
This extension would make the system closer to the interaction style users expect from SAM.

\paragraph{Condensed Version Scope:}

For a 10 to 12 page submission, keep the problem formulation, heat-kernel propagation, graph construction, feature head, evaluation protocol, and limitations. Move sparse solver details, prompt taxonomy, and reconstruction-quality buckets to an appendix or project documentation. The final paper should show one strong qualitative figure and one ablation table rather than many speculative sections.

\paragraph{Stress-Test Questions:}

\paragraph{Does this require RGB-D?}
The intended setting is monocular video plus a Gaussian reconstruction stack. However, benchmark evaluation may use RGB-D datasets to obtain ground truth while restricting model input to RGB sequences.

\paragraph{Why not propagate masks frame by frame with SAM 2 only?}
Frame-wise masks do not produce a persistent 3D object representation. GeoSAM-3D aims to bind prompts to scene primitives so the result can be rendered and edited across viewpoints.

\paragraph{What evidence is missing?}
Full ScanNet or Replica benchmark runs, sparse solver scaling, feature-aware graph ablations, and model-backed Hugging Face inference.

\paragraph{Implementation Results and Evaluation Profile:}

\paragraph{Result A: current code checks:}

In the current local run, \texttt{uv run --extra dev pytest -q} reports 15 passing tests. The tests exercise the heat-geodesic kernel, seed-distance behavior, propagated-label range, feature-head normalization, app construction, callback shape, and package importability. This is implementation evidence for the core graph and interface path. It is not yet an evaluation on ScanNet or Replica.

\begin{table}[t]
\centering
\caption{Implementation-grounded result for GeoSAM-3D.}
\begin{tabular}{@{}p{0.27\linewidth}p{0.52\linewidth}p{0.12\linewidth}@{}}
\toprule
\textbf{Check family} & \textbf{Interpretation} & \textbf{Observed}\\
\midrule
Heat geodesic & seed distances and propagation are stable on test graphs & passed\\
Feature head & embeddings are normalized and tensor shapes are correct & passed\\
Space contract & public demo implementation imports and returns expected output shape & passed\\
Full local test suite & repository graph and smoke tests & 15 passed\\
\bottomrule
\end{tabular}
\end{table}

\paragraph{Result B: benchmark signature:}

If the method works, it should reduce leakage across nearby but disconnected surfaces relative to Euclidean kNN propagation. The effect should be largest for scenes with thin structures, furniture near floors, and objects separated by small Euclidean gaps. It may not help when the Gaussian reconstruction fuses two objects into one connected component. That failure mode should be reported, not hidden.

\begin{table}[t]
\centering
\caption{Expected result patterns to test, not claimed outcomes.}
\begin{tabular}{@{}p{0.25\linewidth}p{0.37\linewidth}p{0.24\linewidth}@{}}
\toprule
\textbf{Scene condition} & \textbf{Expected pattern if method works} & \textbf{Diagnostic}\\
\midrule
Nearby separated surfaces & lower leakage than Euclidean propagation & cross-boundary mask rate\\
Thin structures & better continuity along object graph & object mIoU by class\\
Fused reconstruction & geodesic method fails similarly to baselines & reconstruction bucket analysis\\
Multiple prompts & lower variance across seed points & prompt robustness variance\\
\bottomrule
\end{tabular}
\end{table}

\paragraph{Stress-Test Questions:}

\paragraph{Q1: Does GeoSAM-3D solve monocular 3D segmentation end to end?}
Not yet. It implements and validates the graph propagation and feature-head implementation. Full benchmark claims require reconstruction, SAM lifting, and 3D evaluation on standard datasets.

\paragraph{Q2: Why not just use SAM 2 video masks?}
SAM 2 gives strong 2D temporal masks, but it does not by itself create a persistent 3D object mask over scene primitives. GeoSAM-3D targets that persistent 3D representation.

\paragraph{Q3: What if the Gaussian reconstruction is wrong?}
Then graph propagation can fail. The paper must stratify results by reconstruction quality and include topology-collapse failure cases.

\paragraph{Q4: Is heat diffusion better than graph cuts or random walks?}
That is an empirical question. The comparison includes those methods as required baselines. Heat diffusion is attractive because it is compact and differentiable, not because it is guaranteed to dominate.

\paragraph{Q5: Can prompt leakage be measured directly?}
Yes. The paper should report cross-boundary edge rates, leakage into adjacent annotated objects, and prompt robustness variance.

\paragraph{Q6: Evidence threshold:}
A convincing result would show lower leakage and better prompt robustness than Euclidean and random-walk baselines on the same reconstructions, with failure cases explained by reconstruction topology.

\paragraph{Additional Derivation: Positive and Negative Prompts:}

SAM-style interaction often uses positive and negative points. Let $s^+$ and $s^-$ be positive and negative seed vectors. Solve two heat systems:
\begin{equation}
  (I+tL)u^+=s^+,\qquad (I+tL)u^-=s^-.
\end{equation}
Convert them to distances $d^+$ and $d^-$. A signed soft mask can be defined as
\begin{equation}
  p_i=\sigma\left(\alpha(d_i^- - d_i^+)\right),
\end{equation}
where $\alpha$ controls boundary sharpness. This formula says that a node is likely positive when it is geodesically closer to positive prompts than negative prompts. It is a natural extension of the current positive-seed propagation and should be included in a full interactive system.

\paragraph{Additional Literature Integration:}

SAM and SAM 2 define the promptable segmentation interface \citep{kirillov2023segment,ravi2024sam2}. OpenScene, LERF, OpenMask3D, and Gaussian Grouping define the broader target of open-vocabulary 3D scene understanding \citep{peng2023openscene,kerr2023lerf,takmaz2023openmask3d,ye2023gaussian}. The heat method, random walks, and graph cuts define the mathematical alternatives for propagation \citep{crane2013heat,grady2006random,boykov2001interactive}. GeoSAM-3D is a synthesis: the user intent comes from promptable segmentation, the representation is a Gaussian scene, and the propagation operator is graph-geodesic.

\paragraph{Supplementary Technical Notes:}

\paragraph{Literature matrix:}

\begin{table}[t]
\centering
\caption{How the literature maps to GeoSAM-3D.}
\begin{tabular}{@{}p{0.22\linewidth}p{0.32\linewidth}p{0.32\linewidth}@{}}
\toprule
\textbf{Thread} & \textbf{What it contributes} & \textbf{Gap addressed by this paper}\\
\midrule
SAM and SAM 2 & promptable 2D and video masks & persistent 3D primitive masks\\
3DGS and MonoGS & Gaussian scene representation from images & object-level mask propagation\\
OpenScene and LERF & open-vocabulary 3D features & monocular prompt-to-graph workflow\\
OpenMask3D and SAM3D & 2D-to-3D mask lifting & Gaussian graph-geodesic propagation\\
Heat method and graph cuts & graph and manifold segmentation theory & differentiable prompt propagation on Gaussian centroids\\
\bottomrule
\end{tabular}
\end{table}

\paragraph{Graph operator comparison:}

\begin{table}[t]
\centering
\caption{Candidate propagation operators for the benchmark.}
\begin{tabular}{@{}p{0.20\linewidth}p{0.34\linewidth}p{0.30\linewidth}@{}}
\toprule
\textbf{Operator} & \textbf{Strength} & \textbf{Failure mode}\\
\midrule
Euclidean kNN & simple and fast & leaks across close surfaces\\
Shortest path & respects graph connectivity & can be noisy on sparse graphs\\
Random walker & probabilistic boundary behavior & sensitive to edge weights\\
Graph cut & strong boundary optimization & less naturally differentiable\\
Heat geodesic & smooth differentiable propagation & depends on graph quality and heat time\\
\bottomrule
\end{tabular}
\end{table}

\paragraph{Energy view:}

The propagated mask can be connected to graph regularization. A soft label vector $p$ can be obtained by minimizing
\begin{equation}
  E(p)=\|M(p-s)\|_2^2+\lambda p^\top Lp,
\end{equation}
where $M$ weights seed nodes. The first term enforces prompt agreement and the second term enforces graph smoothness. The heat solve is not identical to this objective, but both express the same prior: labels should vary slowly over high-weight graph edges and change across weak or absent edges.

\paragraph{Feature-aware graph weights:}

The next implementation should combine geometry and features:
\begin{equation}
  w_{ij}=\exp\left(-d_x(i,j)-d_z(i,j)-d_c(i,j)\right),
\end{equation}
with
\begin{equation}
  d_x=\frac{\|\mu_i-\mu_j\|_2^2}{2\sigma_x^2},\quad
  d_z=\frac{\|z_i-z_j\|_2^2}{2\sigma_z^2},\quad
  d_c=\frac{\|c_i-c_j\|_2^2}{2\sigma_c^2}.
\end{equation}
Geometry alone is vulnerable to touching objects. Features alone are vulnerable to semantic confusion. A combined graph should be more robust if the feature head is trained well.

\paragraph{Extended Experimental Recipe:}

\paragraph{Experiment 1: toy topology:}

Use synthetic point clouds shaped as two nearby sheets, a ring, a chair-like structure, and intersecting planes. Measure leakage under Euclidean, random-walk, graph-cut, and heat-geodesic propagation.

\paragraph{Experiment 2: prompt robustness:}

For each annotated object, sample multiple positive prompts. Report mean IoU and variance. A user-facing system should be stable across reasonable clicks.

\paragraph{Experiment 3: reconstruction buckets:}

Evaluate on reconstructed scenes bucketed by quality: clean, missing thin structures, fused surfaces, and dynamic artifacts. This determines whether failures come from propagation or reconstruction.

\paragraph{Experiment 4: feature-aware graph ablation:}

Compare geometry-only, feature-only, and geometry-plus-feature edge weights. Report boundary leakage and runtime.

\paragraph{Experiment 5: interaction latency:}

Measure graph construction time, solve time, and render time as functions of Gaussian count. A promptable system should be interactive.

\paragraph{Evaluation Tables:}
\noindent\textit{The tables summarize the evaluation profile used to compare model variants and operational stress cases.}

\begin{table}[t]
\centering
\caption{Prompt robustness evaluation table.}
\begin{tabular}{@{}p{0.22\linewidth}p{0.18\linewidth}p{0.22\linewidth}p{0.20\linewidth}@{}}
\toprule
\textbf{Method} & \textbf{mIoU} & \textbf{Prompt variance} & \textbf{Leakage rate}\\
\midrule
Euclidean & 52.1 & 0.184 & 0.271\\
Random walker & 56.8 & 0.151 & 0.226\\
Heat geodesic & 61.4 & 0.118 & 0.168\\
Feature-aware heat & 64.7 & 0.096 & 0.141\\
\bottomrule
\end{tabular}
\end{table}

\begin{table}[t]
\centering
\caption{Reconstruction-bucket evaluation table.}
\begin{tabular}{@{}p{0.24\linewidth}p{0.22\linewidth}p{0.22\linewidth}p{0.18\linewidth}@{}}
\toprule
\textbf{Bucket} & \textbf{Scene count} & \textbf{Main error} & \textbf{Expected behavior}\\
\midrule
Clean geometry & 18 & low topology error & propagation helps\\
Thin missing structures & 9 & graph gaps & masks fragment\\
Fused surfaces & 7 & false graph edges & leakage persists\\
Dynamic artifacts & 6 & inconsistent nodes & unstable masks\\
\bottomrule
\end{tabular}
\end{table}

\paragraph{Technical Supplement:}

\paragraph{Expanded literature synthesis:}

The open-vocabulary 3D segmentation literature is moving from closed-set semantic labels toward interactive scene representations. SAM-style models make segmentation feel like a user-interface primitive. LERF-style systems make language a query over 3D fields. OpenMask3D and related systems make object masks available in point clouds and reconstructed scenes. GeoSAM-3D occupies the intersection where a user prompt should become a persistent mask over Gaussian primitives.

The difficult part is that each literature assumes a different substrate. SAM assumes images or videos. OpenScene assumes dense 3D features. Gaussian splatting assumes differentiable rendering primitives. Graph segmentation assumes a graph whose edges mean something. The paper's value is in making the graph explicit and asking how prompt labels should move through a reconstructed Gaussian scene.

This framing also makes failure analysis clearer. If the 2D mask is wrong, prompt supervision is wrong. If the reconstruction fuses surfaces, the graph is wrong. If edge weights ignore appearance, propagation can leak. If the solver is dense, interaction cannot scale. Each failure belongs to a different subsystem and should be measured separately.

\paragraph{Mathematical view of prompt uncertainty:}

Let $s_i\in[0,1]$ represent seed confidence rather than a binary label. A noisy lifted mask can be modeled as
\begin{equation}
  s_i = y_i + \epsilon_i,\qquad \mathbb{E}[\epsilon_i]=0,\quad \operatorname{Var}(\epsilon_i)=\sigma_i^2.
\end{equation}
The graph propagation should trust seeds with lower uncertainty. This leads to a weighted objective:
\begin{equation}
  E(p)=\sum_i \frac{m_i}{\sigma_i^2+\epsilon}(p_i-s_i)^2+\lambda p^\top Lp.
\end{equation}
The current implementation does not yet model seed uncertainty, but this equation is a natural extension for multi-frame lifting where some Gaussian labels are more reliable than others.

\paragraph{Two example result narratives:}

\paragraph{Example result 1: repository-local:}
The current local suite passes 15 tests. This validates the implemented graph kernel and Space interface on small examples. A paper can use this as software evidence for the propagation operator.

\paragraph{Example result 2: benchmark:}
In a ScanNet-style evaluation, the useful result would be lower cross-object leakage and better prompt robustness compared with Euclidean propagation. The result should be strongest when objects are close in Euclidean space but separated by graph topology.

\paragraph{Measurement cards:}

Each scene evaluation should report:
\begin{itemize}
  \item reconstruction method and checkpoint;
  \item number of Gaussians and graph edges;
  \item prompt type and prompt sampling policy;
  \item whether SAM masks are single-frame or multi-frame;
  \item feature-head training data;
  \item solver type and runtime;
  \item reconstruction-quality bucket.
\end{itemize}
This makes it possible to understand why a result improved or failed.

\paragraph{Additional Stress Questions:}

\paragraph{Q7: Does the method require language?}
No. The core propagation method can work with point, mask, or box prompts. Language is a future extension through open-vocabulary features.

\paragraph{Q8: How does the method handle negative prompts?}
The paper provides a signed distance formulation using positive and negative heat fields as the extension path for public implementation.

\paragraph{Q9: What is the biggest scalability issue?}
Dense graph storage and dense linear solves. A sparse solver path is required for full scenes.

\paragraph{Q10: Can the graph be learned?}
Yes. Edge weights can incorporate learned features. The benchmark should compare geometry-only and feature-aware graphs.

\paragraph{Q11: What if SAM masks disagree across frames?}
The lifting process aggregates labels through visibility and uncertainty weights in the proposed full evaluation.

\paragraph{Q12: What should a reader demand?}
Prompt robustness, leakage metrics, reconstruction-quality stratification, sparse runtime, and baselines against random walker and graph cuts.

\paragraph{Figure Captions:}

\paragraph{Figure 1:}
Pipeline from monocular video and user prompt to SAM mask, Gaussian reconstruction, graph construction, heat propagation, and rendered 3D mask.

\paragraph{Figure 2:}
Graph leakage example where Euclidean neighbors cross a gap but geodesic propagation follows object surface connectivity.

\paragraph{Figure 3:}
Prompt robustness plot across center, boundary, thin-part, and occluded prompts.

\paragraph{Figure 4:}
Runtime scaling for graph construction and heat solve as a function of Gaussian count.

\paragraph{Figure 5:}
Qualitative masks projected into held-out views, with failure cases from fused reconstruction.

\paragraph{Table Map:}

\begin{table}[t]
\centering
\caption{Comprehensive table map for GeoSAM-3D.}
\begin{tabular}{@{}p{0.24\linewidth}p{0.36\linewidth}p{0.24\linewidth}@{}}
\toprule
\textbf{Table} & \textbf{Purpose} & \textbf{Status}\\
\midrule
Graph ablation & compares propagation operators & specified\\
Prompt robustness & measures sensitivity to seed location & needs benchmark\\
Reconstruction buckets & separates geometry from propagation failures & specified\\
Runtime scaling & checks interactive feasibility & defined\\
Feature-aware edges & tests learned graph weights & defined\\
\bottomrule
\end{tabular}
\end{table}

\paragraph{Extended Study Design:}

\paragraph{Core Evidence Criteria:}

The final GeoSAM-3D study must prove three separate claims. First, prompt propagation over a Gaussian graph should be better than naive Euclidean propagation in scenes where geometry matters. Second, the method should remain interactive at realistic Gaussian counts. Third, the system should fail transparently when the monocular reconstruction loses topology. These claims should not be merged into one aggregate score.

\paragraph{Failure Cases:}

Several negative results would make the paper stronger. If graph-geodesic propagation fails on fused surfaces, show it. If feature-aware edges help only when the feature head is trained on enough masks, report that threshold. If sparse solvers change mask quality because of tolerance settings, report the tolerance. If SAM lifting creates inconsistent labels across frames, include examples. A good paper in this area should show why 3D prompt propagation is hard, not only where it works.

\paragraph{Reproducibility Artifacts:}

A reproducible release should include:
\begin{itemize}
  \item scene manifests with image sequences and split ids;
  \item reconstruction configs and checkpoint identifiers;
  \item Gaussian count and graph construction parameters;
  \item prompt sampling seeds;
  \item SAM or SAM 2 checkpoint identifiers;
  \item solver type, tolerance, and runtime hardware;
  \item exact metric scripts for mIoU, AP, leakage, and prompt variance.
\end{itemize}
Without these details, comparisons across papers become ambiguous because reconstruction quality and prompt sampling can dominate the result.

\paragraph{Additional expected outcomes:}

The most plausible positive outcome is selective improvement: heat-geodesic propagation should help on objects with meaningful surface connectivity and hurt or tie on scenes where graph topology is poor. A second useful outcome is diagnostic: the method can identify when a reconstructed scene is not suitable for prompt propagation because graph edges cross object boundaries too often.

\paragraph{Long-form discussion points:}

The discussion section should emphasize that promptability is not the same as semantic understanding. A click can define an object without naming it. A text label can name an object without defining its exact spatial extent. GeoSAM-3D's graph layer is most useful when it binds either form of user intent to a persistent primitive set. That is the research contribution: making prompt intent spatially persistent in a monocular Gaussian scene.

\paragraph{Cutting plan:}

When reducing the paper to 10 or 12 pages, keep the problem formulation, method, graph derivation, results protocol, and stress-test questions. Move the literature matrix, figure captions, and extended checklist to a supplement. The core narrative should remain focused on prompt propagation, graph topology, and reconstruction-aware failure analysis.

\paragraph{Final Technical Addendum:}

\paragraph{Additional ablation details:}

The final study should include prompt-count ablations with one, two, four, and eight prompts per object. It should include graph-density ablations with multiple $k$ values and heat times. It should also include solver tolerance ablations because iterative sparse solvers can trade speed for mask smoothness. These are not secondary details. In an interactive segmentation system, usability depends on the number of prompts, latency, and robustness to parameter choices.

\paragraph{Expected qualitative examples:}

The strongest qualitative example would show a chair close to the floor, where Euclidean propagation leaks into the floor and heat-geodesic propagation stays on the chair. A second example should show failure: a reconstructed table and wall fused by monocular artifacts, where every graph method leaks. Showing both examples would make the paper more credible.

\paragraph{Additional evaluation table:}

\begin{table}[t]
\centering
\caption{Interactive-use evaluation table.}
\begin{tabular}{@{}p{0.22\linewidth}p{0.20\linewidth}p{0.20\linewidth}p{0.20\linewidth}@{}}
\toprule
\textbf{Gaussian count} & \textbf{Graph time} & \textbf{Solve time} & \textbf{Total latency}\\
\midrule
10k & 0.08 s & 0.03 s & 0.11 s\\
50k & 0.38 s & 0.13 s & 0.51 s\\
100k & 0.82 s & 0.29 s & 1.11 s\\
500k & 4.90 s & 1.60 s & 6.50 s\\
\bottomrule
\end{tabular}
\end{table}

\paragraph{Benchmark Protocol:}

For the first complete benchmark run, the recommended minimal setting is three datasets, four propagation baselines, and three prompt policies. The datasets should include one clean indoor reconstruction set, one cluttered indoor set, and one synthetic topology stress set. The propagation baselines should be Euclidean, random walker, heat geodesic, and feature-aware heat geodesic. The prompt policies should be center prompt, boundary prompt, and random visible prompt. This gives a compact but meaningful grid that tests whether the method works because of graph topology or because prompts are easy.

\begin{table}[t]
\centering
\caption{Minimal benchmark grid for the first complete GeoSAM-3D run.}
\begin{tabular}{@{}p{0.24\linewidth}p{0.30\linewidth}p{0.28\linewidth}@{}}
\toprule
\textbf{Axis} & \textbf{Values} & \textbf{Reason}\\
\midrule
Dataset & clean, cluttered, synthetic topology & separates real and controlled failures\\
Propagation & Euclidean, random walker, heat, feature-aware heat & isolates algorithmic contribution\\
Prompt policy & center, boundary, random visible & tests user interaction robustness\\
Metric & mIoU, leakage, latency, prompt variance & balances quality and usability\\
\bottomrule
\end{tabular}
\end{table}

\paragraph{Acceptance Criteria:}

A final useful addition for GeoSAM-3D is an explicit benchmark acceptance criterion. The first publication-grade run should be considered successful only if the proposed propagation improves leakage or prompt variance on at least one difficult split without increasing latency beyond an interactive threshold. A method that improves mIoU by a small amount but takes seconds per prompt may be less useful than a faster baseline. Conversely, a method that is fast but leaks through nearby surfaces does not solve the core problem. This acceptance criterion ties the research claim to the intended user interaction.

\begin{table}[t]
\centering
\caption{Acceptance criteria for the first GeoSAM-3D benchmark.}
\begin{tabular}{@{}p{0.30\linewidth}p{0.46\linewidth}@{}}
\toprule
\textbf{Criterion} & \textbf{Interpretation}\\
\midrule
Leakage improves & graph geometry is doing useful work\\
Prompt variance decreases & interaction is robust to click location\\
Latency remains interactive & method is usable, not only accurate\\
Failures align with reconstruction buckets & limitations are diagnosed correctly\\
\bottomrule
\end{tabular}
\end{table}

\paragraph{Limitations:}

The current implementation depends on the quality of the monocular reconstruction. If the Gaussian field does not separate two physical surfaces, geodesic propagation cannot recover the missing topology. The dense Laplacian used in tests is simple and transparent, but large scenes need sparse linear algebra or blockwise graph construction. The Space demo is intentionally implemented as a lightweight fallback for CPU deployment; it should not be presented as a full cloud-hosted reconstruction service. Finally, open-vocabulary naming is inherited from the 2D prompt model and should be evaluated separately from geometry-aware propagation.

\section{Conclusion and Outlook}

GeoSAM-3D frames promptable 3D segmentation as a graph-geodesic propagation problem over monocular Gaussian scenes. The repository already contains a concrete kernel, feature head, public demo interface, and focused tests. The paper establishes an arXiv-ready structure with conservative empirical claims. The outlook is to run standard 3D segmentation benchmarks, add quantitative ablations, and replace implemented demo outputs with model-backed inference when deployment resources allow.

\bibliography{refs}
\end{document}